\documentclass{article}
\usepackage{spconf,amsmath,graphicx}
\usepackage{multicol}
\usepackage{makecell}
\usepackage{stfloats}

\title{Learning Discriminative Representations for Fine-Grained Diabetic Retinopathy Grading}

\name{Li Tian$^1$ \qquad Liyan Ma$^{\star1,3}$ \qquad Zhijie Wen$^{\star2}$ \qquad Shaorong Xie$^{1}$\qquad Yupeng Xu$^{4}$\thanks{* The corresponding author \{liyanma,wenzhijie\}@shu.edu.cn.    This research is supported by the National Natural Science Foundation of China (61991410, 61991415, 11701357, 11971296).}}
  
  \address{$^{1}$School of Computer Engineering and Science, Shanghai University, Shanghai, China \\
   $^{2}$Department of Mathematics, College of Sciences, Shanghai University, Shanghai, China \\
   $^3$Shanghai Institute of Intelligent Science and Technology, Tongji University,Shanghai, China, \\
   $^4$Department of Ophthalmology, Shanghai General Hospital, National Clinical Research Center for Eye\\Diseases
  Shanghai Jiao Tong University School of Medicine, Shanghai Key Laboratory of Ocular Fundus\\Disease, Shanghai Engineering Center for Visual Science and Photomedicine, Shanghai, 200080, China.}
  
\begin{document}

\maketitle

\begin{abstract}
  Diabetic retinopathy (DR) is one of the leading causes of blindness. However, no specific symptoms of early DR lead to a delayed diagnosis, which results in disease progression in patients. To determine the disease severity levels, ophthalmologists need to focus on the discriminative parts of the fundus images. In recent years, deep learning has achieved great success in medical image analysis. However, most works directly employ algorithms based on convolutional neural networks (CNNs), which ignore the fact that the difference among classes is subtle and gradual. Hence, we consider automatic image grading of DR as a fine-grained classification task, and construct a bilinear model to identify the pathologically discriminative areas. In order to leverage the ordinal information among classes, we use an ordinal regression method to obtain the soft labels. In addition, other than only using a categorical loss to train our network, we also introduce the metric loss to learn a more discriminative feature space. Experimental results demonstrate the superior performance of the proposed method on two public IDRiD and DeepDR datasets.
\end{abstract}


\section{Introduction}

\label{sec:intro}

Diabetic retinopathy(DR) is one of the most common vision-threatening diseases. Early diagnosis and timely treatment are vital for preventing visual impairment and have been proved to be able to reduce the risk of blindness by up to 90 percent. However, there are no obvious symptoms of retinopathy in the early stage, which requires diabetic patients to frequently take the fundus image examination to avoid vision problems. The diagnosis of DR is a highly time-consuming task since there are lots of types of lesions with diverse features. Therefore, an automatic and effective image grading algorithm for DR is in urgent need.

With the great success of deep learning technology, computer-aided diagnosis (CAD) plays an important role in the medical diagnosis. Recent methods \cite{gulshan2016development,pratt2016convolutional,gargeya2017automated}  based on CNNs have significantly improved the performance of DR classification. The accurate diagnosis still exists great challenges: 1) Be different from the coarse-grained classification, DR grading is a fine-grained classification task with large intra-class variance. 2) Medical image labeling is a costly task. 3) The datasets follow a long-tailed class distribution. In this paper, we propose a fine-grained diabetic retinopathy grading scheme to address the above challenges. 
\begin{figure*}[!t]
  \begin{minipage}[b]{1.0\linewidth}
    \centering
    \centerline{\includegraphics[width=17cm]{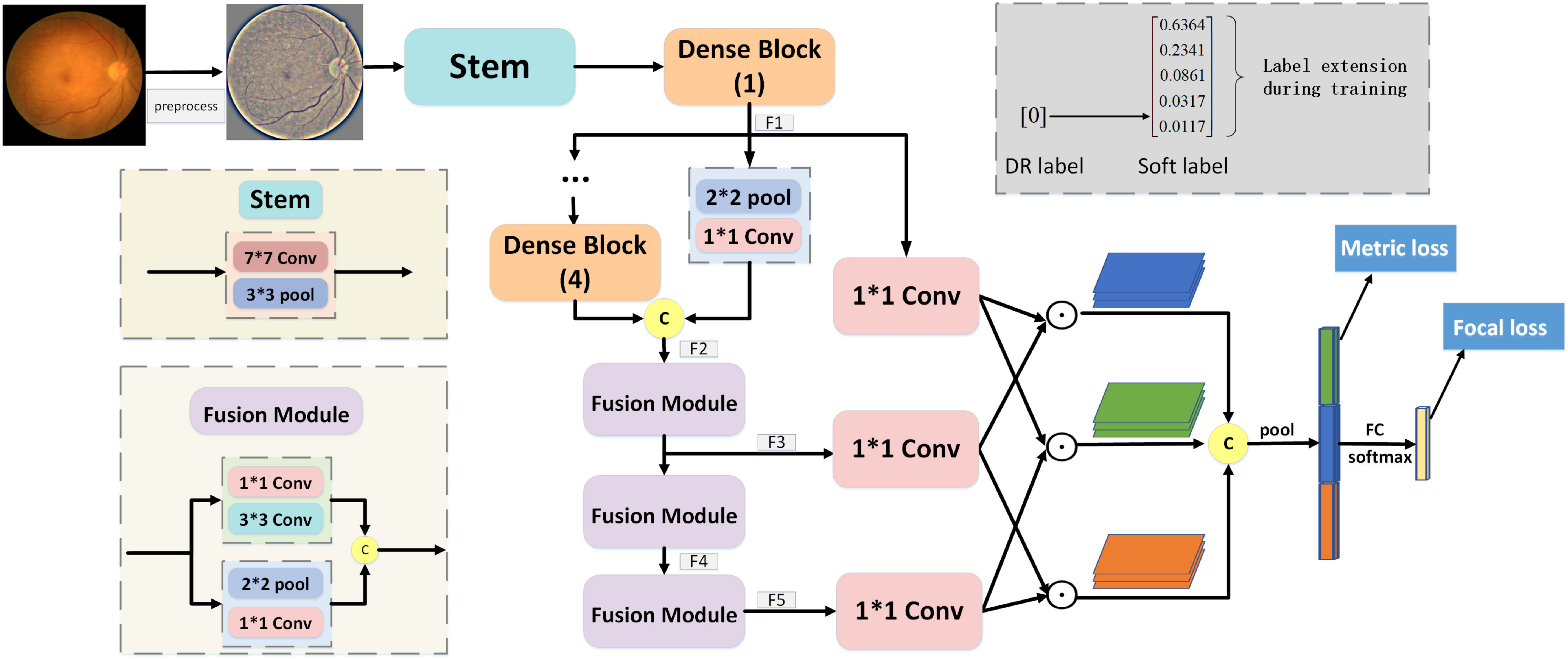}}
  \end{minipage}
  \caption{Illustration of the proposed network architecture. $\bigodot$ represents the element-wise multiplication.}
  \vspace{-0.5cm}  
\end{figure*}
Image recognition task includes coarse-grained image classification and fine-grained image classification. The goal of coarse-grained image classification is to recognize the categories with large appearance differences such as cats, dogs, etc. While concentrating on the fine categories with small inter-class variance, such as the species of birds, fine-grained classification puts more effort into extracting local discriminative features. For DR detection, the disease severity level mostly relies on the small lesion regions, such as microaneurysms, hemorrhages, and so on. Therefore, unlike most previous works, we consider DR grading as a fine-grained classification task. We build our neural network architecture based on the hierarchical bilinear pooling (HBP) \cite{yu2018hierarchical} which has been demonstrated to be effective for fine-grained classification.

Conventional image classification methods encode the ground truth labels into one-hot labels that treat all wrong classes equally. However, most previous works ignore the fact that the severity of DR follows a natural order. It inspires us to utilize the ordinal regression approach \cite{diaz2019soft} to encode the interclass distance. 
In addition, the success of deep learning methods highly depends on the ability of feature representations. Since metric learning tends to learn a feature embedding space where images belonging to the same class get closer than that from other classes.  We combine the effective feature embedding technique  \cite{qian2019softtriple} to learn a more discriminative representation for fundus images.

Our main contributions can be summarized as: (1) Identifying the discriminative parts of fundus images based on the fine-grained approach. (2) Using an ordinal regression method to take advantage of ordinal information. (3) Obtaining a more discriminative embedding space via deep metric learning. (4) Numerous experimental results show the superior performance of the proposed method.
\section{RELATED WORK}
With its remarkable success in medical image analysis in the past decade, deep learning has been applied to the classification of DR. Most works for DR grading directly apply a feature extractor followed by a classifier, which are effective for general image classification task. Gulshan et al. \cite{gulshan2016development} used the Inception network \cite{szegedy2016rethinking} to detect diabetic retinopathy and obtained promising results. Pratt et al. \cite{pratt2016convolutional} proposed a method with a 10-layer of CNNs followed three dense connected layers to make a diagnosis of DR grading in real-time. Gargeya et al. \cite{gargeya2017automated} utilized the residual network \cite{he2016deep} to extract features, and then used these features with other metadata to train the decision tree for classification. However, these methods take the DR classification task as a conventional natural image classification which ignores the small inter-class variance and the interclass ordinal relationships of the DR problem. 

Bajwa et al. \cite{bajwa2019combining} proposed an ensembling method (ResNet and DenseNet \cite{huang2017densely} as two coarse-grained classification methods, NTS-Net \cite{yang2018learning} and SBS layer \cite{recasens2018learning} as two fine-grained classification methods) to predict the severity of DR. Although utilizing the fine-grained classification approach, the ensembling approach of four neural networks is somewhat complicated and time-consuming. Moreover, ordinal information among classes is not taken into account.
\section{METHODOLOGY}
This section presents the proposed method which is illustrated in Fig. 1. To reduce the inﬂuence of lighting conditions, we first preprocess the fundus images. The preprocessed images are fed into the backbone to extract the feature maps. Then, a bilinear module \cite{yu2018hierarchical} captures the cross-layer interaction information. Finally, the classifer gives the predictions.

\subsection{Network architecture}
Because the progression of DR is a gradual process, there are subtle variations among different sub-categories. Moreover, the appearance of fundus images belonging to the same category is quite different since there exit various types of lesions. To tackle these challenges, we develop the network based on the HBP module which is proposed for fine-grained classification.

Each preprocessed fundus image is fed into the CNN based feature extractor to obtain its representation. HBP utilizes VGG16 \cite{simonyan2014very} as the backbone. To better capture the multi-scale information, the proposed method uses DSOD architecture \cite{shen2017dsod} which combines DenseNet169 with a feature pyramid network (FPN) \cite{lin2017feature} as the backbone.  

After obtaining feature maps, the bilinear module will be used to fuse multi-scale features from layers F1, F3, F5 by concatenating the outputs of bilinear vectors as follows:
$$Z=P^T concat(U^Tf_1 \odot  V^Tf_3,U^Tf_1 \odot S^Tf_5,V^Tf_3\odot S^Tf_5)$$
where $f_1,f_3,f_5 $ are the output features from layers F1, F3, F5, $U, V, S$ are projection matrices which map the features into bilinear vectors and $P$ is the classification parameter matrix. 

\subsection{Loss fuction}
Be different from the traditional classification problems, there exists the ordinal information among disease severity levels in most medical datasets. Therefore, we use an ordinal regression approach to obtain a soft label \cite{diaz2019soft} rather than the classical one-hot label which does not contain ordinal information.  In particular, for i-th example $x_i$, the ground-truth label $y_i$ is written as $(y_{i,1},y_{i,2},...,y_{i,C})^T$ and 
$$y_{i,j}=\frac{e^{-\phi(r_{i,t},r_{i,j})}}{\sum_{c=1}^Ce^{-\phi(r_{i,t},r_{i,c})}}\ ,$$ 
where C is the number of class, $r_{i,t}$ is the t-th rank of true metric value. For example, when there do not exist any abnormalities in fundus images, the value of $r_{i,t}$ is 0. Rank $r_{i,j}$ is one of the C 
ordinal lables, and $\phi(r_{i,t},r_{i,j})$ is the pre-defined metric function to penalize the distance between $r_{i,t}$ and $r_{i,j}$. For the penalty function $\phi$, we use the simple squared error.

In order to learn a discriminative feature space, we use a metric loss \cite{qian2019softtriple} as:
$$L_{metric}=-log\frac{exp(\lambda(S'_{i,y_i}-\delta))}{exp(\lambda(S'_{i,y_i}-\delta))+\sum_{j\neq y_i}exp(\lambda S'_{i,j})}\ ,$$
and $$S'_{i,y_i}=\sum_{k=1}^K\frac{exp(\frac{1}{\gamma}x^T_iw^k_c)}{\sum_{k=1}^Kexp(\frac{1}{\gamma}x^T_iw^k_c)}x^T_iw^k_c \ ,$$
where $\lambda$ is a smooth term. For each class $c$, there are K centers for the large intra-class variance. $\delta$ is a pre-defined margin and $w$ is the parameter of the fully connected layer. $S'_{i,y_i}$ is the relaxed similarity between $x_i$ and the class $c$. For each category with a representative center, SoftMax loss with a smooth term $\lambda$ is equivalent to a smoothed triplet loss. While in real-world data,
 each class contains multiple centers due to the large intra-class variance. By expanding the weight matrix of each class to have multiple columns, this metric loss can be optimized with triplets. 

Since there exists a serious class imbalance challenge, we use the focal loss \cite{lin2017focal} to deal with the imbalance problem of DR datasets:
$$L_{focal} = -(1-p_t)^{\gamma}log(p_t),$$
where $\gamma$ is a focusing parameter and $p_t$ is the prediction probability of label $y$. In the experiments, we set $\gamma$ to be 2.

The final loss is the combination of focal loss and metric loss:
$$L = \alpha L_{metric}+\beta L_{focal},$$ 
where $\alpha $ and $\beta$ are the hyperparameters.
\section{Experiments}
\label{sec:pagestyle}
\subsection{Datasets and evaluation metrics}
\label{sec:format}
In this paper, we use 5-fold cross-validation on two public DR datasets, namely IDRiD and DeepDR, which are provided by the IEEE International Symposium on Biomedical Imaging (ISBI) in 2018 and 2020 respectively. According to the clinical International Clinical Diabetic Retinopathy Disease Severity Scale \cite{wilkinson2003proposed}, the progression of DR from healthy to proliferative phase can be developed into five categories: 1) healthy, 2) mild-NPDR, 3) moderate-NPDR, 4) severe-NPDR and 5) PDR. PDR and NPDR stand for proliferative diabetic retinopathy and nonproliferative diabetic retinopathy respectively. The IDRiD and DeepDR datasets contain 516 and 1600 images with resolutions of 4288 x 2848 and 1736 x 1824 respectively. 

In order to evaluate the proposed method on the IDRiD and DeepDR datasets. We utilize the quadratic weighted kappa which typically ranges from 0 to 1. This metric is used to measure the consistency between two ratings. The larger the value is, the higher the consistency is. We also present the accuracy of each class by calculating the confusion matrix. 
\subsection{Implementation details}
\label{sec:format}
The quality of some fundus images is poor due to the illumination factors. To reduce the influence of lighting conditions, we preprocess the datasets with the method of Ben's preprocessing$\footnote{https://www.kaggle.com/ratthachat/aptos-eye-preprocessing-in-diabetic-retinopathy}$. Besides, we remove the redundant black background. Then, we resize the images to be 512x512 pixels.

During the training phase, the data augmentations we used include random resized crop, random affine, random horizontal, and vertical flip. 

Our experiments are conducted on four NVIDIA GTX 1080Ti GPUs based on the Pytorch. All backbone networks are pre-trained on the imageNet dataset. For the optimizer, we choose Stochastic Gradient Descent (SGD) and 
the momentum is set to be 0.9. Batch size and epoch are set to be 32 and 150 respectively. The learning rate is initially set to be 0.01 and divided by 10 every 50 epochs. $\alpha$ and $\beta$ are set to be 0.5.
\subsection{Comparison with the state-of-the-art }
\label{sec:format}
In this section, we compare the performance of the proposed method with the state-of-the-art fine-grained classification methods, such as B-CNN \cite{lin2015bilinear}, HBP \cite{yu2018hierarchical}, DFL \cite{wang2018learning}, and PMG \cite{du2020fine}. Meanwhile, the proposed method is also compared with the ensembling method \cite{porwal2020idrid} which is the fusion of AlexNet and GoogleNet. As illustrated in Table 1, our method achieves superior performance. This shows the effectiveness of the proposed approach.
\begin{table}[t]
  \footnotesize
  \caption{The average quadratic weighted kappa of the proposed approach with the state-of-the-art methods. AG\_Net represents the ensemble model of AlexNet and GoogleNet.}
  \hspace*{\fill} \\
  \centering
  \begin{tabular}{l c c c}\hline
    Methods&Backbone&IDRiD&DeepDR\\\hline
    B-CNN (ICCV15) \cite{lin2015bilinear}&VGG16&0.8631&0.8702\\
    HBP (ECCV18) \cite{yu2018hierarchical}&VGG16&0.8511&0.8586\\
    DFL (CVPR18) \cite{wang2018learning}&ResNet50&0.8804&0.8926\\
    PMG (CVPR20) \cite{du2020fine}&ResNet50&0.8694&0.8825\\\hline    
    AG\_Net (MIA20) \cite{porwal2020idrid}&AlexNet$+$GoogleNet&0.8573&0.8644\\\hline
    Our Method&DSOD&\bf{0.8874}&\bf{0.9050}\\\hline
  \end{tabular}
  \label{tab:Margin_settings}
\end{table}
\vspace{-0.3cm}
\subsection{Ablation study}
\label{sec:format}
This section presents the effectiveness of the components in the proposed method. Firstly, replacing the backbone of the HBP model with DSOD obtains better results as shown
in Table 2. In addition, the focal loss also has some impacts.
The compared results presented in rows 3-4 of Table 2 indicate that the ordinal regression approach can improve the average quadratic kappa score by at least 1\% on both datasets. This performance suggests the importance of the ordinal information in the DR problem.
\begin{figure}[htb]
  \begin{minipage}[b]{0.32\linewidth}
    \centering
    \centerline{\includegraphics[width=2.7cm]{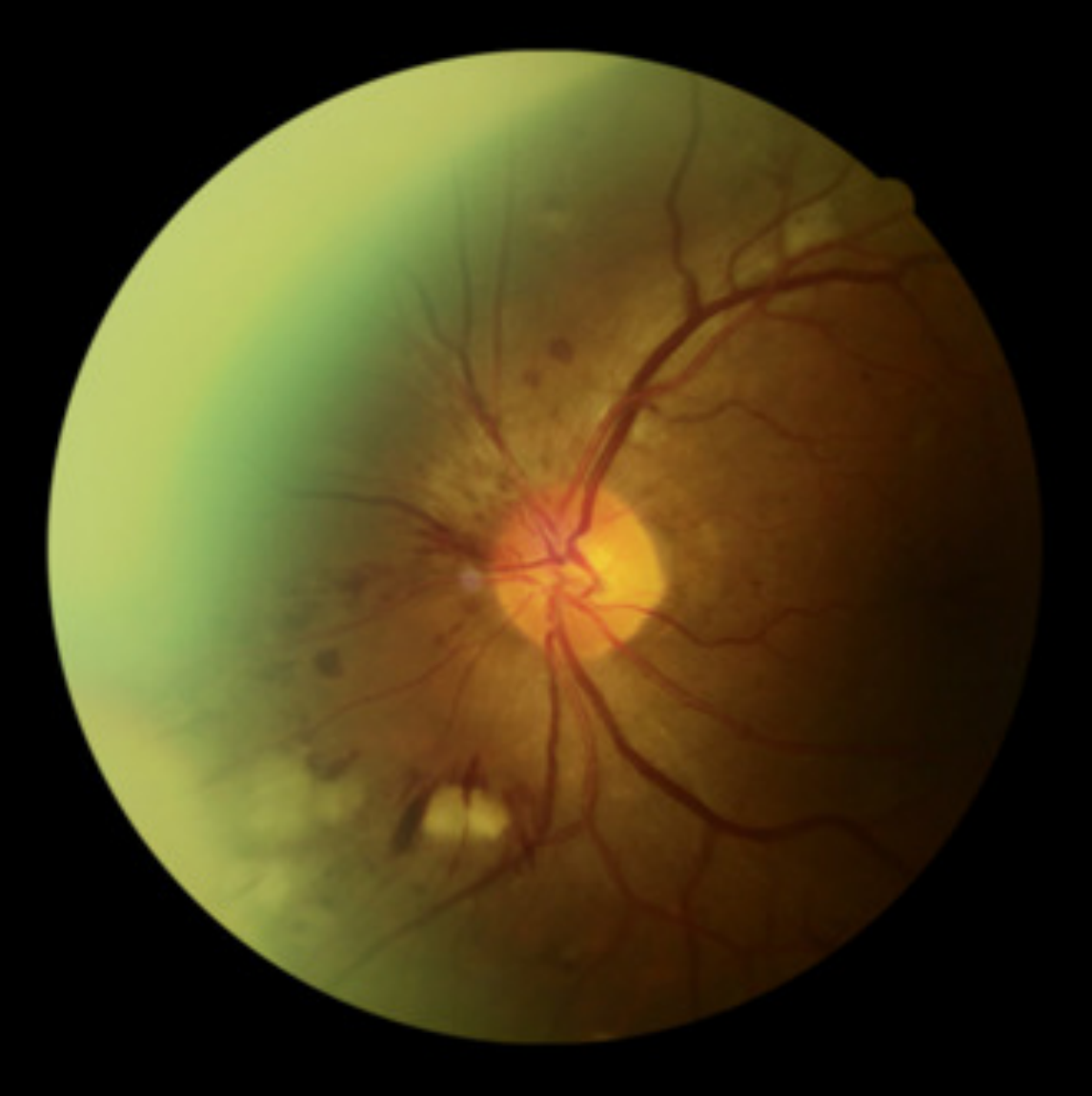}}
    \centerline{(a)}\medskip
  \end{minipage}
  \begin{minipage}[b]{0.33\linewidth}
    \centering
    \centerline{\includegraphics[width=2.7cm]{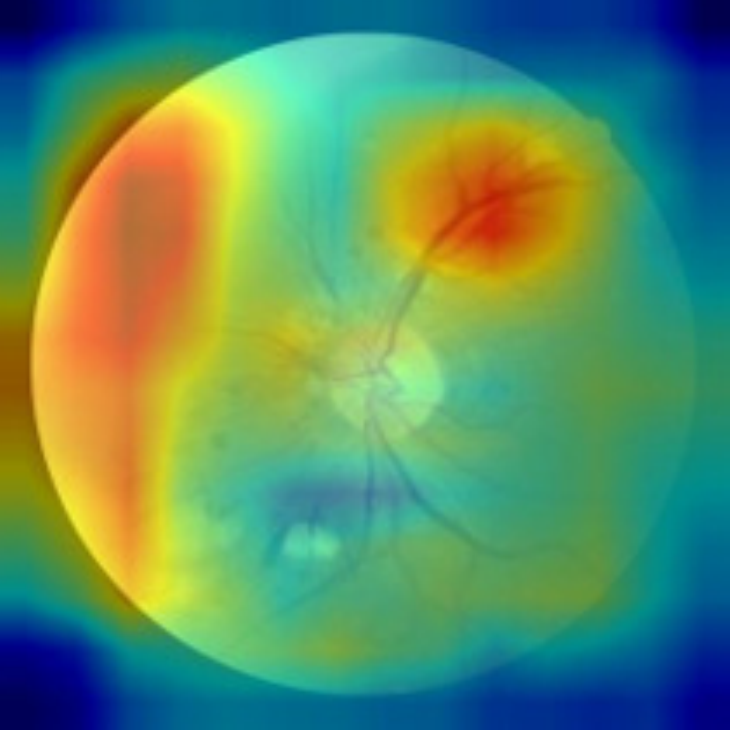}}
    \centerline{(b)}\medskip
  \end{minipage}
  \hfill
  \begin{minipage}[b]{0.32\linewidth}
    \centering
    \centerline{\includegraphics[width=2.7cm]{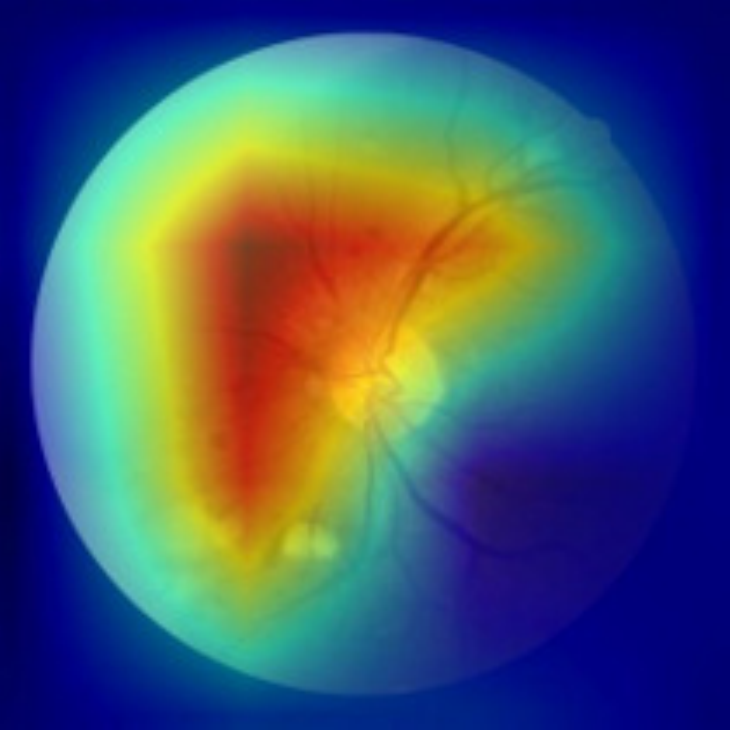}}
    \centerline{(c)}\medskip
  \end{minipage}
  \vspace{-0.3cm}
  \caption{Class activation maps. (a): an example from DeepDR dataset; (b): the visualization of the HBP model; (c): the visualization of the proposed model.}
  \label{fig:res}
  \end{figure}
\vspace{-0.2cm}
The introduction of a metric loss can further improve the average quadratic weighted kappa by around 0.7\% and 1.0\% on IDRiD and DeepDR datasets respectively as shown in the last two rows of Table 2. It means the metric loss could help to learn a better feature embedding.
\subsection{Visualization}
\label{sec:format}
To illustrate the superior performance of the proposed method, we conduct visualization experiments via Grad-CAM \cite{selvaraju2017grad}. Column (a)-(c) in Fig.2 give an original DR image sampled from the DeepDR dataset, the activation maps from HBP and the proposed method respectively. Our method focuses on the discriminative regions rather than the black background and less informative regions. The class activation maps show that the proposed method truly makes predictions based on the discriminative parts.
Furthermore, to better present the accuracy of each class, we plot the confusion matrix based on the DeepDR dataset as shown in Fig. 3. The diagonal elements of the confusion matrix represent the accuracy of each category. The phenomenon that those wrong predictions are close to the ground truth labels is extremely important for DR diagnosis. Due to the serious imbalance of the DeepDR dataset, only 92 images for PDR, the accuracy of PDR is slightly worse.
\begin{table}[t]
  \small
  \centering
  \caption{Ablation studies for the proposed method. OR represents the ordinal regression approach. ML is the metric loss. The performance is evaluated by the average quadratic weighted kappa.}
  \hspace*{\fill} \\
  \begin{tabular}{l c c c}\hline
    Methods&Backbone&IDRiD&DeepDR\\\hline
    HBP&VGG&0.8511&0.8586\\
    HBP (CEloss)&DSOD&0.8643&0.8826\\
    HBP (FLloss)&DSOD&0.8681&0.8848\\
    HBP$+$OR (FLloss)&DSOD&0.8805&0.8983\\
    HBP$+$OR (FLloss$+$ML)&DSOD&0.8874&0.9050\\\hline 
  \end{tabular}  
  \label{tab:Margin_settings}
\vspace{-0.05cm}
\end{table}
\begin{figure}[htb]
  \setlength{\abovecaptionskip}{0.cm}
  \setlength{\belowcaptionskip}{-0.cm}
  \begin{minipage}[b]{1.0\linewidth}
    \centering
    \setlength{\abovecaptionskip}{0.cm}
    \centerline{\includegraphics[width=0.85\textwidth]{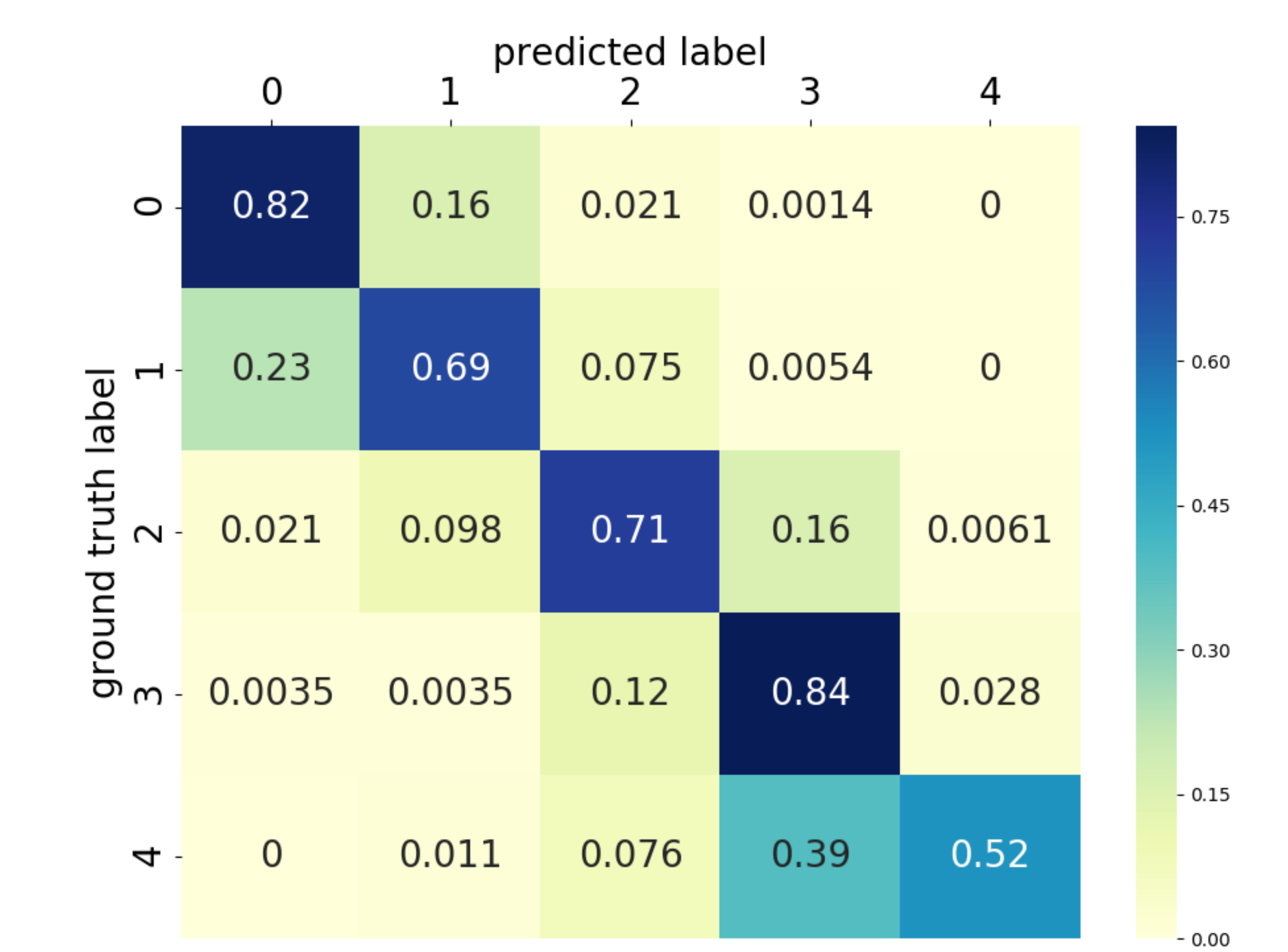}}
  \end{minipage}
  \caption{Confusion matrix of the proposed approach on DeepDR dataset.}
  \label{fig:res}
  \end{figure}
\vspace{-0.8cm}
\section{Conclusion}
\label{sec:typestyle}
To get the discriminative feature representation for fundus images, we proposed a method based on the fine-grained classification framework. With the consideration of ordinal information among classes, the proposed method achieves competitive performance. Furthermore, the metric loss pushes semantically similar examples closer than that from different classes in the feature embedding space. Therefore, the proposed method achieves state-of-the-art performance on two public DR datasets which are more challenging than general images.
\bibliographystyle{IEEEbib}
\bibliography{reference,refs}  
\end{document}